\documentclass[conference]{IEEEtran}
\IEEEoverridecommandlockouts
\usepackage{cite}
\usepackage{amsmath,amssymb,amsfonts}
\usepackage{algorithmic}
\usepackage{algorithm}
\usepackage{graphicx}
\usepackage{textcomp}
\usepackage{xcolor}
\usepackage{booktabs}
\usepackage{graphicx}
\usepackage{caption}
\usepackage{float} 
\usepackage{subcaption}
\usepackage{array}
\usepackage[margin=54pt]{geometry}
\def\BibTeX{{\rm B\kern-.05em{\sc i\kern-.025em b}\kern-.08em
    T\kern-.1667em\lower.7ex\hbox{E}\kern-.125emX}}
\begin{document}

\title{Exploring Global and Local Information for Anomaly Detection with Normal Samples}

\author{\IEEEauthorblockN{Fan Xu}
\IEEEauthorblockA{\textit{University of Science and Technology of China} \\
Hefei, China \\
82920}
\and
\IEEEauthorblockN{Nan Wang$^*$}
\IEEEauthorblockA{\textit{Beijing Jiaotong University} \\
Beijing, China \\
80442}
\and
\IEEEauthorblockN{Xibin Zhao}
\IEEEauthorblockA{\textit{Tsinghua University} \\
Beijing, China \\
82927}
}

\maketitle

\begin{abstract}
Anomaly detection aims to detect data that do not conform to regular patterns, and such data is also called outliers. The anomalies to be detected are often tiny in proportion, containing crucial information, and are suitable for application scenes like intrusion detection, fraud detection, fault diagnosis, e-commerce platforms, et al. However, in many realistic scenarios, only the samples following normal behavior are observed, while we can hardly obtain any anomaly information. To address such problem, we propose an anomaly detection method GALDetector which is combined of global and local information based on observed normal samples. The proposed method can be divided into a three-stage method. Firstly, the global similar normal scores and the local sparsity scores of unlabeled samples are computed separately. Secondly, potential anomaly samples are separated from the unlabeled samples corresponding to these two scores and corresponding weights are assigned to the selected samples. Finally, a weighted anomaly detector is trained by loads of samples, then the detector is utilized to identify else anomalies. To evaluate the effectiveness of the proposed method, we conducted experiments on three categories of real-world datasets from diverse domains, and experimental results show that our method achieves better performance when compared with other state-of-the-art methods.
\end{abstract}


\section{Introduction}
The purpose of anomaly detection\cite{ref1} is to identify abnormal patterns from the normal behaviors in a data set, and these abnormal patterns are called anomalies or outliers. The applications of anomaly detection are wide-ranging and include areas such as fraud detection\cite{ref2}, wind control systems\cite{ref3}, suspicious transaction monitoring\cite{ref4}, fault detection\cite{ref5}, et al. There are now many advanced anomaly detection algorithms, including supervised, unsupervised, and semi-supervised methods. However, many of these algorithms are difficult to use in practical applications due to the scarcity of abnormal samples in the real world. Furthermore, anomalies often contain multiple patterns, and new patterns may emerge continuously, making it difficult to learn from partially observed anomalies.

While, in many realistic scenarios, researchers can hardly get anomalous samples but can simply obtain a batch of normal samples. For example, in network traffic logs\cite{ref16}, network traffic attacks are rough to identify and are often complex and changeable. In addition, anomalous attacks that have never appeared before may occur. Nevertheless, the logging system can utilize its built-in storage to obtain part of normal traffic information simply. Now the critical problem is that there is no valid solution for taking advantage of normal sample information sensibly.

\begin{figure}[!t]
\centering
\includegraphics[width=3.5in]{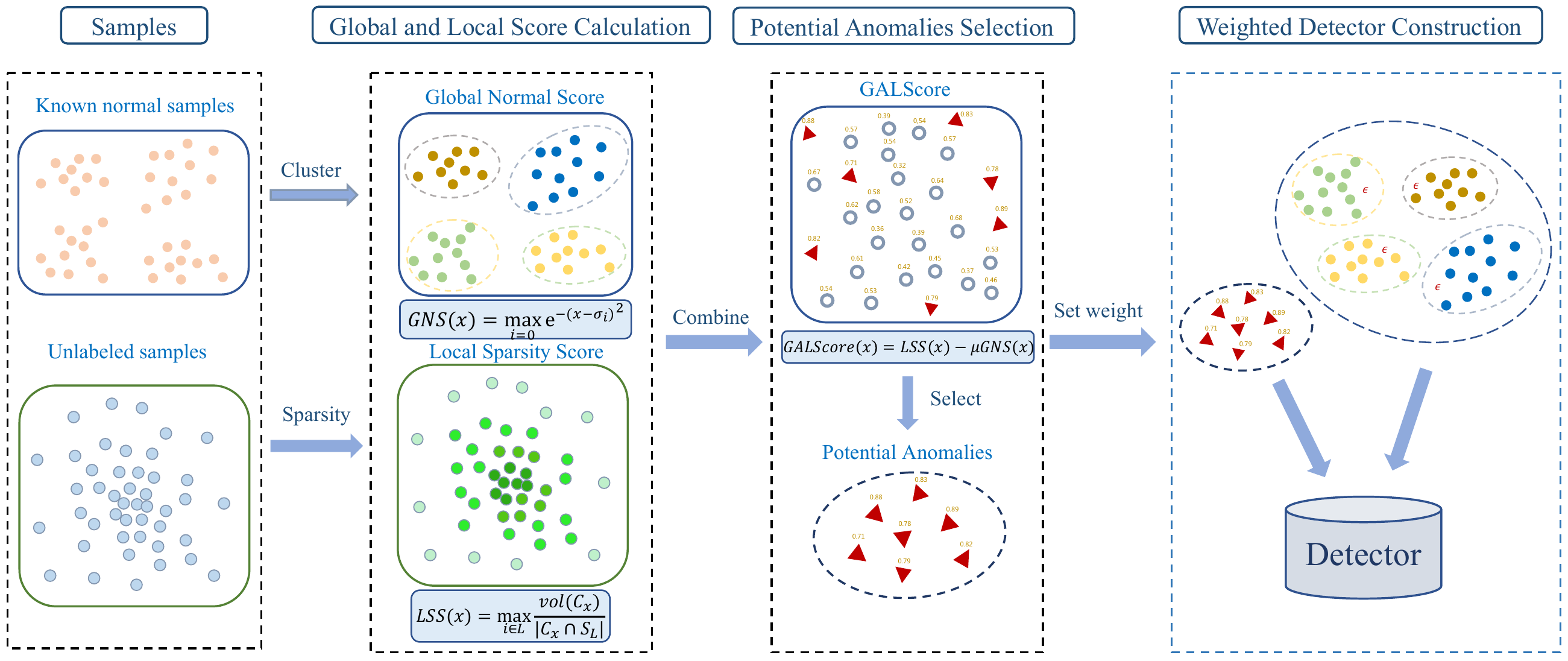}
\caption{The overall framework of GALDetector}
\label{framework}
\end{figure}

To solve the issues, this paper proposes an anomaly detection method GALDetector which only utilizes several observed normal samples and combines the global and local information. The specific motivation is to extract local anomaly information while making full use of the observed normal samples to obtain global normal information. The method can be divided into a three-stage task and the overall procedure is shown in Fig.~\ref{framework}. In the first stage, the observed normal samples are clustered\cite{ref18} to learn different patterns of normal samples. Then we compute the similarity of each sample with its nearest clustering center, which is considered as a global normal score. After that, the local sparsity scores of the samples are calculated using a density-based integration method\cite{ref19}. In the second stage, global normal score of each sample is computed by combining these two scores, and a threshold is set to filter out the probable anomalous samples. Then diverse weights are assigned to all samples according to global normal scores. In the third stage, we build a weighted-based detection model based on weights for training, which in turn identifies anomalies from unlabeled sample set. Experiments on various real-world datasets will demonstrate the effectiveness of our method.

The main contributions of our work are summarized below:
\begin{itemize}
\item Considering that in anomaly detection scenarios, anomaly information is usually difficult to obtain and often requires a lot of human and material resources. Instead our method only uses easily available normal data to complete the anomaly detection task, and is comparable to current state-of-the-art algorithms.
\item We propose an anomaly detection method GALDetector based on normal samples combining both global and local information. By calculating global normal score and local sparsity score of each sample simultaneously, we manage to capture the critical elements of whether a sample is anomalous or not.
\item We conduct extensive experiments on loads of benchmark real-world datasets and the results show that GALDetector outperforms or matches other popular anomaly detection methods.
\end{itemize}


The remainder of this paper is organized as follows. In section II, we introduce related work on anomaly detection in recent years. In section III, we demonstrate our proposed method GALDetector step by step. In section IV, we demonstrate the experimental setup and the corresponding results of our proposed method compared with other state-of-the-art methods. Then we finally conclude this paper in section V.

\section{Related Work}

Anomaly detection task aims to identify points that does not conform to what is normal or expected in a data set. Anomaly detection has been employed in many different scenarios, mainly in cyber security, industrial control, e-commerce platforms\cite{ref24}, finance, et al. Loads of algorithms have been applied to deal with this task, and they can be classified into unsupervised, supervised, and semi-supervised anomaly detection algorithms according to what kind and how much label information we can utilize. 

Among the developed methods, supervised anomaly detection methods assume that sufficient labeled data are available and that these methods can derive useful information from labeled data. Supervised-SVM\cite{ref25} uses hyperplane in multi-dimensional space to divide data points into classes. KNN\cite{ref34} calculates the average distance of k-nearest neighbors to identify anomalies. AutoEncoder\cite{ref26} is based on the use of artificial neural networks that encode the data by compressing it into the lower dimensions and then decode it to reconstruct the original input.

While labeled data are not available in all scenarios, lots of current works on anomaly detection are focused on unsupervised methods. For example, LOF\cite{ref32} is a typical local density-based anomaly detection algorithm; iForest\cite{ref33} is a random forest based algorithm, which utilizes the distance of leaf nodes from the root node in each tree to portray the anomalous degree; PCA\cite{ref35} is a typical dimension reduction-based algorithm, it projects the samples into a low-dimensional space and calculates the deviation in all directions to calculate the anomaly scores. These methods can be broadly employed since there is no need for any label information.

However, in many application scenarios, it is burdensome to achieve expected performance employing unsupervised methods. Sometimes we can obtain some real labels of the data, so semi-supervised methods are researched. In the scenarios where part of anomalous samples can be known in the initial stage, while no normal sample is labeled, PU learning\cite{ref36} is explored to solve such problem, and they can be mainly divided into three kinds. The traditional PU learning can be regarded as a two-stage task, in the first step we select reliable normal samples from unlabeled samples, and in the second step we train the classifier based on known abnormal samples and reliable normal samples. Biased learning\cite{ref37} treats the set of unlabeled samples as a set of normal samples with noise and assigns relatively high and low regularization parameters to known abnormal and normal samples. Class Prior Incorporation\cite{ref38} assigns weights to positive and negative samples according to the proportion of positive and negative samples. It then builds a logistic regression model to fit the weighted samples. Nevertheless, anomalies cannot be simply classified into one class, and ADOA\cite{ref39} is proposed to tackle such problem. ADOA differs from PU learning which treats all anomalies as one conceptual center. However, ADOA clusters anomalies into k clusters, then calculates the degree of isolation and similarity with the nearest anomalous clustering center simultaneously. Later ADOA constructs a weighted multi-classification model to detect anomalies. 

Moreover, in several scenarios, we can only obtain several normal samples without any anomaly information in the initial stage, and some methods mostly similar with unsupervised or supervised methods are explored to solve such setting. For instance, SVDD\cite{ref48} uses kernel function to map the data to a high-dimensional space, and the smallest possible hypersphere is found to enclose the normal data. GMM\cite{ref49} models a Gaussian mixture model on normal data, estimating parameters with maximum likelihood. Nevertheless, there is no mature method that can integrate global and local information of labeled normal samples and unlabeled samples, such method will be explored next in this paper.

\section{Anomaly Detector With Global and Loacal Information}

\subsection{Overview and Preliminaries}

In this section, we introduce the proposed Anomaly Detector With Global and Loacal Information algorithm(GALDetector). Fig.~\ref{framework} illustrates the overall framework of the proposed method, which consists of three main stages, i.e., global and local score calculation, potential anomalies selection and weighted detector construction.

In the first stage, we will calculate both local sparsity scores and global normal scores. For local sparsity scores, a local density integration algorithm based on partial identification is utilized to calculate the local sparsity scores of all unlabeled samples. And for global normal scores, a clustering operation is used to identify multiple patterns of normal samples and set them into different clusters, then the global normal scores of all samples are calculated based on these clustering centers. In the second stage, we select potential anomalies selection and set weights to all the samples. After the global normal scores of the samples are calculated according to the global normal scores and the local sparsity scores computed above, we set threshold to select the potential anomalies from the unlabeled samples. Then we assign corresponding weights to known normal samples and selected potentially anomalous samples based on the global normal scores. In the third stage, a weighted detector construction is built based on previous weights and the model parameters are optimized by training such samples. Our model will be employed to identify anomalous on the remaining unlabeled samples.

We let $\kappa = R^{d}$ denotes the sample space and $y=\left \{ 0,1 \right \}$ represents label space, in which $y=0$ is for normal samples and $y=1$ is for anomalous samples. The corresponding data set we selected is $D=\left \{ (x_{1},y_{1}),...,(x_{p},y_{p}),x_{p+1},...,x_{N} \right \} $. Among all given $N$ samples, the first $p$ samples are known normal ones $D_{n}=\left \{ (x_{1},y_{1}),...,(x_{p},y_{p}) \right \}$, and the leftover $(N-p)$ samples are unlabeled ones $D_{u}=\left \{ x_{p+1},x_{p+2},...,x_{N} \right \}$.

\subsection{Global and Local Score Calculation}

\textbf{Local Sparsity Score(LSS).} The local sparsity score of each sample is first calculated. We adopt a certain strategy to divide the high-dimensional attribute space into independent subcubes with maximized sparsity variance, so as to portray the local sparsity characteristics of the nodes. To be specifically, we utilize a integration algorithm originated from decision trees, and each tree is built by a partial sample of the data set. Then each tree partitions the high-dimensional sample space into various dense and sparse subcubes.

For each tree, we firstly normalize each coordinate to $[0,1]$ and select a random sample $S\in \Omega $ of $N$ points. From the root, we pick a coordinate $j \in [d]$ at each internal node, then we select $f_{1}<f_{2}<\cdots< f_{p-1}$ which partition the corresponding coordinate $I_{j}$ into $p$ intervals, and split the current node space into $p$ intervals. The number of partitions $p$ is a hyper-parameter, and utilizing $p=5$ works well in practice. Then we partition the property space by selecting coordinates and breakpoints continuously, and the operation stops until reaching the maximum depth or only one point existing in a subcube. Finally the whole sample space is divided into several subcubes and the volume of subcube C can be represented as the product of the interval's length on each selected coordinate:
\begin{equation}
    vol(C)= {\textstyle \prod_{j}^{}}len(I_{j}).
    \small
\end{equation}

Then the overall sparsity of C could be expressed as :
\begin{equation}
    \rho (S ,C)=\frac{vol(C)}{\left | C\cap S \right |},
    \small
\end{equation}
in which $\left | C\cap S \right |$ demonstrates the number of points in the intersection of subcube $C$ and the selected sample set $S$. The local sparsity score of point $x$ is displayed as $LSS(x)$, which is the mean sparsity of subcube containing $x$ in all trees and we have $T$ trees in total.
\begin{equation}
    LSS(x)=\frac{1}{T} \sum_{i=1}^{T}\rho (S_{i},C_{x})=\frac{1}{T} \sum_{i=1}^{T} \frac{vol(C_{x})}{\left|C_{x}\cap S_{i} \right|}.
    \small
\end{equation}

Our intention is to split C into sparse and dense subcubes, so it is intuitively that we ought to maximize the variance in the sparsity. Maximizing the variance has the advantage that it is equivalent to a well-studied histogram problem\cite{ref41} and allows for a very efficient streaming algorithm\cite{ref42}. Through such algorithm we can compute the best split along each coordinate.

\renewcommand{\algorithmicrequire}{\textbf{Input:}}  
\renewcommand{\algorithmicensure}{\textbf{Output:}}
\begin{algorithm}[t]
    \scriptsize
    \caption{Calculation of GALScore}\label{alg}
    \begin{algorithmic}
        \REQUIRE{All samples $D$, Normal samples $D_{n}$, Unlabeled samples $D_{u}$, Number of trees $L$, Number of samples $N$, Max degree $m$, Max depth $h$, Number of clusters $n$}
        \ENSURE{$GALScore Calculation$}
        \FOR {each $i \in [1,L]$}
        \STATE Create root node $v$
        \STATE Let $C(v)=[0,1]^{d}$, $P(v)\subseteq S$ be a random subset of size N.
        \STATE For ${j \in [d]}$, compute the best split into k intervals.
        \STATE Pick $j$ that maximizes variance, split $C$ along $j$ into ${C^{i}}$.
        \STATE For $i \in [k]$ create child $v_{i}$ s.t. $C(v_{i})=C^{i}, P(v_{i})=P(v)\cap C^{i}$.
        \STATE If $depth(v_{i}) \le h$ and $|P(v_{i})|>1$ then $Split(v_{i})$.
        \STATE Else, set $LSS(x) = \max_{i\in L}\frac{vol(C_{x})}{\left|C_{x}\cap S_{L} \right|}$.
        \ENDFOR
        
        \STATE Cluster normal samples $D_{n}$ into $n$ clusters and $\sigma_{r}$ is the $r$-th cluster center
        \FOR {each $x \in D$}
        \STATE $d_{x} = inf$
        \FOR {each $r \in [1,n]$}
        \STATE $d_{x,r} = dist(x, \sigma_{r})$
        \STATE $d_{x} = min(d_{x},d_{x,r})$
        \ENDFOR
        \STATE $GNS(x) = e^{-d_{x}^2}  $
        \ENDFOR
        \STATE $GALScore(x) = LSS(x) - \mu GNS(x)$
        \STATE Output $GALScore$.
    \end{algorithmic}
\end{algorithm}

\textbf{Global Normal Score(GNS).} To calculate the global normal score, we need to cluster the known normal samples so as to learn the corresponding normal patterns. Afterwards, by measuring the gap between unlabeled samples and normal patterns, we can obtain a representation of the global normal fraction. For the known normal samples $D_{n}= \left \{ n_{1},n_{2},...,n_{p} \right \}$, we cluster them into k clusters, and each clustering center represents a normal pattern. Many clustering methods are tried here, considering the effectiveness and time complexity, we choose k-means clustering. We use the Euclidean distance to measure, so that the distance between two points is $dist(x_{m},x_{n})=\sqrt{\sum_{j=1}^{d}(x_{mj}-x_{nj})^{2}} $. The k-means algorithm constrains and optimizes the clustering centers by the following equation.

\vspace{-0.3cm}
\begin{equation}
    E = \sum_{i=1}^{k}\sum_{x\in C_{i}}^{}dist(x-\sigma_{i} ),  
    \small
\end{equation}
in which $\sigma_{i}$ is the $i$-th clustering center from the previous stage of clustering. Intuitively, the closer a sample is to its nearest clustering center, the more probably it is to subordinate to a normal pattern. Therefore we use the distance between the unlabeled sample and the nearest normal clustering center to calculate the global normal score. The specific calculation formula of global normal score is as follows:

\vspace{-0.3cm}
\begin{equation}
    \setlength{\abovedisplayskip}{3pt}
    \setlength{\belowdisplayskip}{3pt}
    GNS(x) = \max_{i=1}^{k}e^{-(x-\sigma _{i} )^2}.
    \small
\end{equation}

\subsection{Potential Anomalies Selection and Weights Setting}

\textbf{Global and Local Score(GALScore).} In stage two, we combine the local sparsity score and the global normal score after the two scores are calculated separately, so that we can obtain the anomaly characteristics of both global and local information of the sample points. The whole calculation process is shown in Algorithm~\ref{alg}.

\vspace{-0.3cm}
\begin{equation}
    \setlength{\abovedisplayskip}{3pt}
    \setlength{\belowdisplayskip}{3pt}
    GALScore(x) = LSS(x) - \mu GNS(x),
    \small
\end{equation}
in which $\mu \in R^{+}$ is a hyperparameter to consider the importance of the two scores together. Since the two scores indicate a tendency towards normal and anomalous respectively, therefore the calculation should take the negative sign. In addition, we normalize the combined score GALScore to adjust to the assignment of the weights in the following. 

To facilitate the subsequent model training, we need to select several potential anomalies from the unlabeled samples. The strategy employed here is to pick the highest scores from the combined score by a certain percentage $\delta$ as potential anomalous samples and $\delta $ is often selected from 5\% to 10\%.

Then we set corresponding weights for all observed normal samples and potential anomalies. For normal samples, fixed weights are assigned to them, for instance 0.5. For unlabeled samples, the larger the GALScore is, the more likely it is to be an anomaly. Therefore, the following strategy is used to assign weights to the picked potential anomalous samples.
\begin{equation}
    w(x)=\frac{GALScore(x)}{max_{x} GALScore(x)}. 
    \small
\end{equation}
Thereby, the weights of potential anomalous samples are assigned between [0,1]. As for known normal samples, the imbalance in the number of known normal samples and potential anomalies is taken into account, and the weights of the known normal samples are set uniformly to $\epsilon  \in [0,1]$. Our motivation is assigning larger weights to potential anomalous samples compared to normal samples, so as to solve the problem of unbalanced sample distribution.

\subsection{Weighted Detector Construction}

In stage three, we build and train a weighted binary detector construction using known normal samples and selected potential anomalous samples when distinct weights are assigned to them. The following is the objective function to be optimized.
\begin{equation}
    \sum_{i}^{N}w_{i}l(y_{i},f(x_{i}))+\lambda R(w),
    \small
\end{equation}
in which $w_{i}$ denotes the weight of the sample $x_{i}$, $R(w)$ is the regularization term, and $l(y_{i},f(x_{i}))$ is the loss term. Here we use a commonly employed classifier XGB\cite{ref44}.

\section{Experiments}
In this section, we introduce the testing datasets, contrast algorithms, evaluation criteria, experimental setup and results. All the experiments are conducted with platform of CPU i7 3.2GHZ and 16G RAM. 

\subsection{Datasets}
To evaluate our method, we conduct experiments on three categories of datasets from different domains, containing different percentage of anomalies, diverse data volume, and various dimensions.

\begin{table}[h]
\caption{Details of real-world datasets.}\label{data}
\centering
\renewcommand\arraystretch{1}  
\tabcolsep=2mm    
\begin{tabular}{llll}
\toprule
Data set    & nodes   & dimension   & \#anomalies(\%)         \\
\midrule
Thyroid     & 7200    & 6   & 534(7.42\%)          \\ 
Mammography & 11183   & 6   & 250(2.32\%)          \\ 
Seismic     & 2584    & 15  & 170(6.5\%)           \\ 
Satimage-2  & 5803    & 36  & 71(1.2\%)            \\ 
Vowels      & 1456    & 12  & 50(3.4\%)            \\ 
Musk        & 3062    & 166 & 97(3.2\%)            \\ 
smtp        & 95156   & 3   & 30(0.03\%)           \\ 
http        & 567479  & 3   & 2211(0.4\%)          \\ 
\bottomrule
\end{tabular}
\end{table}

The first category is classification datasets from the UCI\cite{ref45} and openML repository\cite{ref46}, concluding Thyroid, Seismic and Mammography. These three datasets deal with binary classification tasks, and the percentage of less class samples is around 5\%. The second category is multi-categorical datasets, including Satimage-2, Vowels and Musk. These have been categorized into normal and abnormal classes based on existing work\cite{ref47}, and the less class accounts for 1\% to 4\%. The third category of the datasets is smtp and http, they are  originated from the KDD Cup 1999 network intrusion detection task and their percentages of anomalies are all below 0.5\%.

\subsection{Contrast Algorithms and Experimental Setup}

To evaluate the superiority of our method, we compare GALDetector with several state-of-the-art anomaly detection methods. The details of compared methods are introduced as follows.

\noindent For the unsupervised algorithm, the methods we contrast are shown below:
\begin{enumerate}
\renewcommand{\labelenumi}{(\theenumi)}
    \item Principal Component Analysis\cite{ref35}(PCA). This method projects the samples into a low-dimensional space and calculates the deviation in all directions to calculate the anomaly scores.
	\item Isolation Forest\cite{ref33}(iForest). This method utilizes the distance of leaf nodes from the root node in each tree to portray the anomalous degree.
	\item Partial Identification Forest\cite{ref19}(PIDForest). This method builds density-based forests with largest sparsity variance to calculate anomaly scores.
	\item Local Outlier Factor\cite{ref32}(LOF). This method determine whether the point is an outlier by comparing the density of each point with its neighbor point.
\end{enumerate}
For the supervised algorithm, the methods we contrast are displayed below:
\begin{enumerate}
\renewcommand{\labelenumi}{(\theenumi)}
    \item K-Nearest Neighbour\cite{ref34}(KNN). This method calculates the average distance of k-nearest neighbors to identify anomalies.
	\item Supervised SVM\cite{ref25}(S-SVM). This method uses hyperplane in multi-dimensional space to divide data points into normal and anomalous classes.
\end{enumerate}
For the semi-supervised algorithms, the methods we compare are shown below:
\begin{enumerate}
\renewcommand{\labelenumi}{(\theenumi)}
    \item PU learning\cite{ref36}(PUL). This method selects reliable normal samples from unlabeled ones and trains a binary classifier to detect anomalies.
	\item Class Prior Incorporation\cite{ref38}(CPI). This method builds a logistic regression after setting weights to anomalous and unlabeled samples.
	\item Anomaly Detection with Partially Observed Anomalies\cite{ref39}(ADOA). This method utilizes partial observed anomalies to learn different anomalous patterns and then detect anomalies.
\end{enumerate}

For the unsupervised, supervised and semi-supervised algorithms, we adopt different experimental setups respectively. For the unsupervised algorithm, 80\% of the total sample is taken each time to train the model and predict it, make the recall-precision curve, and take the F1 value at the maximum for effect evaluation. For the supervised algorithm, 80\% of the total samples are taken for training and the remaining 20\% for testing. For the semi-supervised algorithms, where PUL, ADOA, and CPI all select 80\% of the total samples as unlabeled samples and 10\% of the anomalous samples as known anomalies for training. For each algorithm and each dataset, 10 independent experiments are conducted and the mean value of the evaluation metrics is taken as the final criterion.

\subsection{Evaluation Criteria}
Since the sample distribution is hugely unbalanced in the anomaly detection task, the percentage of anomalies is generally tiny and below 5\% in most datasets. So that evaluation criteria like accuracy is not much significant. In this paper, we mainly use the AUC value and $F_{1}$-measure as our evaluation criteria, and the details are demonstrated as follows:
\begin{enumerate}
\renewcommand{\labelenumi}{(\theenumi)}
    \item AUC value measures the area under the ROC curve. The ROC curve plots false positive rate on x-axis and false negative rate on y-axis.
	\item $F_{1}$-measure is a joint measurement to evaluate the prediction performance, which takes both the Recall and Precision into consideration, and the valid formula is $F_{1} = \frac{2\times Recall\times Precision}{Recall+Precision} $. In the equation, Recall is the percentage of correctly detected anomalies compared with all the anomalies, and is calculated by $Recall = \frac{TP}{TP+FN} $. And Precision is the percentage of anomalies correctly classified as anomalies compared with the samples which are classified as anomalies, calculated as $Precision = \frac{TP}{TP+FP} $. By setting different thresholds in the classifier, we can attain corresponding PR curve, then we need to select a specific threshold to obtain the $F_{1}$-measure.
\end{enumerate}

\subsection{Experimental Results}

To evaluate the performance of our proposed method, the experiments are employed on loads of various benchmark datasets originating from different domains. Before experiments are performed, normalization is carried out for each data set on all coordinates. 

We demonstrate that GALDetector outperforms or matches lots of state-of-the-art anomaly detection algorithms on various real-world benchmarks. The comparison of the performance of our method with other algorithms was demonstrated in the two tables below. Table~\ref{AUC-result} shows the best AUC that can be obtained by varying the parameter settings, and Table~\ref{F1-result} displays the best $F_{1}$-measure by setting different thresholds simultaneously. As we can see from the two tables, GALDetector is the top performing or joint top performing algorithm in 4 out of the 8 datasets in terms of AUC value, and is the top performing or jointly top performing algorithm in 4 out of the 8 datasets in terms of $F_{1}$-measure respectively.

As is shown in the experiments, our proposed method GALDetector has more stable results on most of the datasets instead of only achieving excellent results on several datasets as other methods do. Compared with the state-of-the-art unsupervised anomaly detection methods, i.e., PCA, iForest, PIDForest and LOF, our proposed method achieves better performance on almost all evaluation metrics for all datasets. More specifically, GALDetector achieves gains of 7.03\%, 2.59\%, 1.13\% and 25.51\% on these four unsupervised methods in terms of overall AUC value respectively. Then in terms of overall $F_{1}$-measure, the improvements are 19.08\%, 15.35\%, 14.99\% and 42.91\% on them separately. This is because unsupervised anomaly detection methods do not use label information, and different methods tend to have distinct effects on different datasets.

\begin{table}[t]
\scriptsize
\caption{Results of AUC value on real-world datasets. We bold the algorithm(s) with the best AUC value.}\label{AUC-result}
\centering
\renewcommand\arraystretch{1.2}  
\tabcolsep=0.12mm    
\begin{tabular}{c|cccc|cc|ccc|c}
\hline
                 &\multicolumn{4}{c|}{Unsupervised}  &\multicolumn{2}{c|}{Supervised}
                 &\multicolumn{3}{c|}{Semi-supervised}   & \multicolumn{1}{c}{Our}  \\ \hline
                 & PCA  & iForest  & PIDForest  & LOF & KNN & S-SVM & ADOA  & PUL & CPI & GALDetector  \\ \hline
Thyroid         & 0.673 & 0.816 & 0.888 & 0.737 & 0.753 & 0.522 & 0.886 & \textbf{0.959} & 0.745 & 0.873\\ \hline
Mammo.          & 0.886 & 0.862 & 0.854 & 0.721 & 0.836 & 0.852 & 0.866 & 0.855 & \textbf{0.919} & 0.863\\ \hline
Seismic         & 0.682 & 0.698 & 0.729 & 0.544 & 0.732 & 0.715 & 0.732 & 0.724 & 0.621 & \textbf{0.736}\\ \hline
Satimage-2      & 0.977 & 0.995 & 0.988 & 0.542 & 0.936 & \textbf{0.996} & 0.963 & 0.953 & 0.939 & 0.979\\ \hline
Vowels          & 0.606 & 0.723 & 0.755 & 0.943 & \textbf{0.975} & 0.625 & 0.841 & 0.897 & 0.951 & 0.824\\ \hline
Musk            & \textbf{1.000} & \textbf{1.000} & \textbf{1.000} & 0.416 & 0.386 & 0.804 & \textbf{1.000}
                & \textbf{1.000} & 0.981 & \textbf{1.000}\\ \hline
smtp            & 0.822 & 0.906 & 0.921 & 0.905 & 0.895 & 0.789 & 0.902 & 0.866 & 0.847 & \textbf{0.932}\\ \hline
http            & 0.996 & \textbf{0.998} & 0.981 & 0.356 & 0.357 & 0.994 & 0.996 & 0.985 & 0.878 & \textbf{0.998}\\ \hline
\textbf{Overall}& 0.831 & 0.875 & 0.889 & 0.646 & 0.734 & 0.787 & 0.895 & 0.892 & 0.860 & \textbf{0.901} \\ \hline
\end{tabular}
\end{table}

\begin{table}[t]
\scriptsize
\caption{Results of $F_{1}$-measure on real-world datasets. We bold the algorithm(s) with the best $F_{1}$-measure.}\label{F1-result}
\centering
\renewcommand\arraystretch{1.2}  
\tabcolsep=0.12mm    
\begin{tabular}{c|cccc|cc|ccc|c}
\hline
                 &\multicolumn{4}{c|}{Unsupervised}  &\multicolumn{2}{c|}{Supervised}
                 &\multicolumn{3}{c|}{Semi-supervised}   & \multicolumn{1}{c}{Our}  \\ \hline
                 & PCA  & iForest  & PIDForest  & LOF & S-SVM & KNN  & ADOA  & PUL & CPI & GALDetector \\ \hline
Thyroid         & 0.224 & 0.349 & 0.387 & 0.298 & 0.392 & 0.541 & 0.421 & \textbf{0.846} & 0.333 & 0.643\\ \hline
Mammo.          & 0.289 & 0.266 & 0.376 & 0.208 & 0.313 & 0.491 & 0.297 & 0.501 & \textbf{0.579} & 0.473\\ \hline
Seismic         & 0.224 & 0.221 & 0.249 & 0.145 & 0.243 & 0.496 & 0.253 & 0.251 & 0.188 & \textbf{0.554}\\ \hline
Satimage-2        & 0.862 & 0.917 & 0.775 & 0.122 & 0.875 & \textbf{0.949} & 0.776 & 0.838 & 0.808 &        
                  \textbf{0.949}\\ \hline
Vowels          & 0.155 & 0.163 & 0.199 & 0.413 & \textbf{0.734} & 0.356 & 0.287 & 0.512 & 0.687 & 0.427\\ \hline
Musk            & 0.995 & 0.979 & \textbf{1.000} & 0.325 & 0.366 & 0.557 & \textbf{1.000}
                & 0.991 & 0.731 & \textbf{1.000}\\ \hline
smtp            & 0.659 & 0.735 & 0.764 & 0.721 & 0.683 & 0.583 & 0.685 & 0.657 & 0.586 & \textbf{0.825}\\ \hline
http            & 0.918 & \textbf{0.994} & 0.903 & 0.198 & 0.954 & 0.217 & 0.976 & 0.962 & 0.673 & 0.981\\ \hline
\textbf{Overall}& 0.541 & 0.578 & 0.582 & 0.303 & 0.570 & 0.524 & 0.587 & 0.682 & 0.573 & \textbf{0.732} \\ \hline
\end{tabular}
\end{table}

We also contrast our method with supervised anomaly detection algorithms, i.e., Supervised SVM and KNN. The experimental results show that our method outperforms supervised methods to some extent. More explicitly, Our method achieves gains of 22.51\% and 16.69\% on these two supervised methods in terms of overall AUC value respectively. And in terms of overall $F_{1}$-measure, the improvements are 16.15\% and 20.78\% on these two methods simultaneously. It is that we do not simply fit the samples to learn an exact delineation criterion, however we integrate global and local information and select several potential abnormal samples. 

When contrasted with semi-supervised anomaly detection methods, i.e., PUL, CPI and ADOA, GALDetector can get results comparable to these three methods, and GALDetector outperforms these methods on at least five of the eight datasets in terms of AUC value. Different from these methods, GALDetector does not utilize any anomaly information, instead it explores the patterns that normal samples should conform to, which is more suitable for real-world scenarios. Moreover, we can see significant enhancement in terms of overall $F_{1}$-measure, and the improvements are 14.46\%, 4.93\% and 15.84\% on those three methods simultaneously.

\section{Conclusion}
In this paper, we solve the anomaly detection problem occurred in scenarios only containing partial labeled normal samples, this is highly different from the traditional unsupervised and supervised algorithms. Moreover, we do not use any anomaly information, which is different from the popular semi-supervised setting where we can obtain a small amount of labeled anomalies beforehand. 

We propose a method called GALDetector, an anomaly detection algorithm based on observed normal samples, which concentrates on global and local information simultaneously. Our proposed method addresses realistic scenarios in many domains where it is challenging to capture anomalous samples while easier to obtain a batch of normal samples. GALDetector starts from normal samples and takes into account both global and local information of samples in the spatial structure, then the corresponding weighted detector is used for training to identify anomalies. We run experiments on real-world datasets from different domains, and the experimental results demonstrate that our proposed method is well adapted to the current problem scenario and has better performance compared with existing state-of-the-art methods. However, owing to absence of anomalous labels, we have to manually select threshold to identify anomalies. In the future we desire to update an adaptive threshold selection method.

\bibliographystyle{unsrt}
\bibliography{file}

\begin{thebibliography}{10}

\bibitem{ref1}
Hongjun Su, Zhaoyue Wu, and Huihui Zhang.
\newblock Hyperspectral anomaly detection: A survey.
\newblock {\em IEEE Geoscience and Remote Sensing Magazine}, 10(1):64--90,
  2021.

\bibitem{ref2}
Fabrizio Carcillo, Yann-A{\"e}l Le~Borgne, Olivier Caelen, Yacine Kessaci,
  Fr{\'e}d{\'e}ric Obl{\'e}, and Gianluca Bontempi.
\newblock Combining unsupervised and supervised learning in credit card fraud
  detection.
\newblock {\em Information sciences}, 557:317--331, 2021.

\bibitem{ref3}
Roberto Cardenas, Rub{\'e}n Pe{\~n}a, Salvador Alepuz, and Greg Asher.
\newblock Overview of control systems for the operation of dfigs in wind energy
  applications.
\newblock {\em TIE}, 60(7):2776--2798, 2013.

\bibitem{ref4}
Aysha Shabbir, Maryam Shabir, Abdul~Rehman Javed, Chinmay Chakraborty, and
  Muhammad Rizwan.
\newblock Suspicious transaction detection in banking cyber--physical systems.
\newblock {\em Computers and Electrical Engineering}, 97:107596, 2022.

\bibitem{ref5}
Miryam~Elizabeth Villa-P{\'e}rez, Miguel~A Alvarez-Carmona, Octavio
  Loyola-Gonzalez, and Miguel~Angel Medina-P{\'e}rez.
\newblock Semi-supervised anomaly detection algorithms: A comparative summary
  and future research directions.
\newblock {\em Knowledge-Based Systems}, 218:106878, 2021.

\bibitem{ref16}
Yanping Zhang, Yang Xiao, Min Chen, Jingyuan Zhang, and Hongmei Deng.
\newblock A survey of security visualization for computer network logs.
\newblock {\em Security and Communication Networks}, 5(4):404--421, 2012.

\bibitem{ref18}
Aristidis Likas, Nikos Vlassis, and Jakob~J Verbeek.
\newblock The global k-means clustering algorithm.
\newblock {\em Pattern recognition}, 36(2):451--461, 2003.

\bibitem{ref19}
Parikshit Gopalan, Vatsal Sharan, and Udi Wieder.
\newblock Pidforest: anomaly detection via partial identification.
\newblock {\em NeurIPS}, 32, 2019.

\bibitem{ref24}
Resty~Kurnia Jamra, Bayu Anggorojati, Dana~Indra Sensuse, Ryan~Randy Suryono,
  et~al.
\newblock Systematic review of issues and solutions for security in e-commerce.
\newblock In {\em 2020 International Conference on Electrical Engineering and
  Informatics (ICELTICs)}, pages 1--5. IEEE, 2020.

\bibitem{ref25}
Mehdi Hosseinzadeh, Amir~Masoud Rahmani, Bay Vo, Moazam Bidaki, Mohammad
  Masdari, and Mehran Zangakani.
\newblock Improving security using svm-based anomaly detection: issues and
  challenges.
\newblock {\em Soft Computing}, 25:3195--3223, 2021.

\bibitem{ref34}
Miao Xie, Jiankun Hu, Song Han, and Hsiao-Hwa Chen.
\newblock Scalable hypergrid k-nn-based online anomaly detection in wireless
  sensor networks.
\newblock {\em IEEE Transactions on Parallel and Distributed Systems},
  24(8):1661--1670, 2012.

\bibitem{ref26}
Zhaomin Chen, Chai~Kiat Yeo, Bu~Sung Lee, and Chiew~Tong Lau.
\newblock Autoencoder-based network anomaly detection.
\newblock In {\em 2018 WTS}, pages 1--5. IEEE, 2018.

\bibitem{ref32}
Markus~M Breunig, Hans-Peter Kriegel, Raymond~T Ng, and J{\"o}rg Sander.
\newblock Lof: identifying density-based local outliers.
\newblock In {\em ACM SIGMOD}, pages 93--104, 2000.

\bibitem{ref33}
Fei~Tony Liu, Kai~Ming Ting, and Zhi-Hua Zhou.
\newblock Isolation forest.
\newblock In {\em 2008 eighth ieee international conference on data mining},
  pages 413--422. IEEE, 2008.

\bibitem{ref35}
Md~Palash Uddin, Md~Al Mamun, and Md~Ali Hossain.
\newblock Pca-based feature reduction for hyperspectral remote sensing image
  classification.
\newblock {\em IETE Technical Review}, 38(4):377--396, 2021.

\bibitem{ref36}
Ya-Lin Zhang, Longfei Li, Jun Zhou, Xiaolong Li, Yujiang Liu, Yuanchao Zhang,
  and Zhi-Hua Zhou.
\newblock Poster: A pu learning based system for potential malicious url
  detection.
\newblock In {\em ACM SIGSAC}, pages 2599--2601, 2017.

\bibitem{ref37}
Kristen Jaskie and Andreas Spanias.
\newblock Positive and unlabeled learning algorithms and applications: A
  survey.
\newblock In {\em 2019 10th International Conference on Information,
  Intelligence, Systems and Applications (IISA)}, pages 1--8. IEEE, 2019.

\bibitem{ref38}
Wee~Sun Lee and Bing Liu.
\newblock Learning with positive and unlabeled examples using weighted logistic
  regression.
\newblock In {\em ICML}, volume~3, pages 448--455, 2003.

\bibitem{ref39}
Ya-Lin Zhang, Longfei Li, Jun Zhou, Xiaolong Li, and Zhi-Hua Zhou.
\newblock Anomaly detection with partially observed anomalies.
\newblock In {\em WWW}, pages 639--646, 2018.

\bibitem{ref48}
Bo~Liu, Yanshan Xiao, Longbing Cao, Zhifeng Hao, and Feiqi Deng.
\newblock Svdd-based outlier detection on uncertain data.
\newblock {\em KAIS}, 34(3):597--618, 2013.

\bibitem{ref49}
Bo~Zong, Qi~Song, Martin~Renqiang Min, and Cheng.
\newblock Deep autoencoding gaussian mixture model for unsupervised anomaly
  detection.
\newblock In {\em ICLR}, 2018.

\bibitem{ref41}
Joung~Taek Yoon, Byeng~D Youn, Minji Yoo, Yunhan Kim, and Sooho Kim.
\newblock Life-cycle maintenance cost analysis framework considering
  time-dependent false and missed alarms for fault diagnosis.
\newblock {\em Reliability Engineering and System Safety}, 184:181--192, 2019.

\bibitem{ref42}
Qichao Ying, Zhenxing Qian, and Xinpeng Zhang.
\newblock Reversible data hiding with image enhancement using histogram
  shifting.
\newblock {\em IEEE Access}, 7:46506--46521, 2019.

\bibitem{ref44}
Tianqi Chen, Tong He, Michael Benesty, Vadim Khotilovich, Yuan Tang, Hyunsu
  Cho, Kailong Chen, et~al.
\newblock Xgboost: extreme gradient boosting.
\newblock {\em R package version 0.4-2}, 1(4):1--4, 2015.

\bibitem{ref45}
Arthur Asuncion and David Newman.
\newblock Uci machine learning repository, 2007.

\bibitem{ref46}
Joaquin Vanschoren, Jan~N Van~Rijn, Bernd Bischl, and Luis Torgo.
\newblock Openml: networked science in machine learning.
\newblock {\em ACM SIGKDD}, 15(2):49--60, 2014.

\bibitem{ref47}
Charu~C Aggarwal and Saket Sathe.
\newblock Theoretical foundations and algorithms for outlier ensembles.
\newblock {\em Acm sigkdd explorations newsletter}, 17(1):24--47, 2015.

\end{thebibliography}

\end{document}